\documentclass[journal]{IEEEtran}
\usepackage{svg}
\usepackage{makecell}
\usepackage{amsmath,amsfonts}
\usepackage{float}
\usepackage[caption=false, font=footnotesize]{subfig}
\usepackage[dvipsnames]{xcolor}
\usepackage{algorithm}
\usepackage{algpseudocode}
\usepackage{array}
\usepackage{textcomp}
\usepackage{stfloats}
\usepackage{url}
\usepackage{verbatim}
\usepackage{graphicx}
\usepackage{cite}
\usepackage{hyperref}
\usepackage{rotating}
\usepackage{soulpos,color}
\usepackage{cleveref}
\makeatletter
\let\ALG@orig@EndFor\EndFor
\renewcommand{\EndFor}{%
  \def\algorithmicendfor{\colorbox{yellow}{\textbf{end for}}}%
  \ALG@orig@EndFor
}
\makeatother

\begin{document}

\title{Reusing Multimodal Agents: Efficient Evaluation of Sound Designs in Games}


\markboth{Journal of \LaTeX\ Class Files,~Vol.~14, No.~8, August~2021}%
{Shell \MakeLowercase{\textit{et al.}}: A Sample Article Using IEEEtran.cls for IEEE Journals}

\author{
\IEEEauthorblockN{Thai Nguyen Van$^{1,*}$, Ibrahim Khan$^{2}$, Mury Fajar Dewantoro$^{3}$, \\Pratch Suntichaikul$^{3}$, Hai Van Pham$^{4}$, Ruck Thawonmas$^{2}$} 

\IEEEauthorblockA{
$^1$Graduate School of Information Science and Engineering, Ritsumeikan University, Japan\\
$^2$College of Information Science and Engineering, Ritsumeikan University, Japan\\
$^3$Independent Researcher\\
$^4$School of Information and Communications Technology, Hanoi University of Science and Technology, Vietnam\\
\thanks{* Corresponding author: gr0557fv@ed.ritsumei.ac.jp \tt\small} }}
\maketitle

\begin{abstract}
This paper introduces a novel framework for efficiently evaluating in-game sound designs using multimodal learning and reinforcement learning. Sound design significantly influences gameplay and accessibility; however, objective evaluation methods, defined here as automated assessments based on agent performance rather than human ratings, often require retraining agents from scratch for each sound design. Our approach first pre-trains a multimodal encoder that encodes visual and auditory information into a shared latent space. A reinforcement learning agent is then trained using this latent representation. We fine-tune the multimodal encoder for each sound design while keeping the trained policy fixed, allowing us to isolate the effects of sound changes on agent behavior. To evaluate the proposed method, we conducted experiments on games from multiple genres, comparing the performance of blind agents (agents using only sound input) under various sound designs. Experimental results suggest that this approach can detect differences in agent performance caused by degraded or altered sound design. Moreover, we significantly reduce the computational cost compared to retraining the agent from scratch. Experimental results suggest that this approach provides a practical and scalable solution for evaluating in-game sound designs. The source code, data and supplementary document are available \footnote{\url{https://github.com/thainv0212/tog-reusing-agents}}.
\end{abstract}


\section{Introduction}
Sound plays a critical role in games, not only enriching the overall player experience but also conveying vital information for perception, navigation, and interaction \cite{10.1007/978-3-030-89394-1_33, ANDERSEN2021222}. Carefully crafted audio cues draw players into the game world, significantly elevating their enjoyment through heightened immersion \cite{10.1525/jsmg.2024.5.4.34}. While the broader scope of game accessibility includes other sensory modalities, such as haptic technologies \cite{mti8010001} to support deaf players, for those with visual impairments, sound is often the primary or sole channel through which gameplay is experienced \cite{yuan2011game, 10.1145/3429360.3468209, 10.1145/3334480.3381660}. As a result, sound design becomes not merely an aesthetic choice but a fundamental component of accessible gameplay. While its aesthetic and subjective qualities remain essential to the player experience, in this work, we focus on the complementary challenge of objectively assessing how informative a sound design is for gameplay.

In recent years, game AI research has begun to explore agents that operate based on audio input alone, leading to new challenges in evaluating the effectiveness of different sound designs, not just from a human-centered perspective, but also through the lens of AI perception and learning\cite{8848088,9893718,pmlr-v162-poklukar22a}. To address these challenges, prior studies have employed in-game audio for representation learning \cite{10.5555/3398761.3398907,10.5555/3535850.3535995}, reinforcement learning \cite{10.1109/CoG52621.2021.9619096,8848088}, and sound evaluation \cite{9893718,khan2025cog}. However, they are hindered by a major limitation: each new sound design typically requires training a deep reinforcement learning (RL) agent from scratch. This retraining process is computationally intensive and time-consuming, making it impractical for iterative sound design testing and comparison. Moreover, retraining introduces stochasticity into the evaluation, complicating efforts to assess the impact of specific sound modifications fairly.

To address this gap, we propose a novel two-phase evaluation framework that decouples sound feature extraction from policy training. In the first phase, an encoder is trained to extract informative auditory features. In the second phase, only the sound-specific portion of this encoder is fine-tuned for each new design, while the policy network remains fixed. This separation allows new sound designs to be evaluated more efficiently, without retraining the entire AI agent from scratch.

This study's contributions are as follows:
\begin{itemize}
    \item Conceptual shift from retraining to adaptation: We propose a novel evaluation paradigm that treats sound design assessment as a representation learning problem rather than a standard reinforcement learning task. By decoupling sound design sensitivity from the policy network, our framework eliminates the need for expensive ``from scratch'' retraining for each design iteration.

    \item Validation of representation sensitivity: We demonstrate that fine-tuning a multimodal encoder, rather than the entire agent, is sufficient to capture the nuances of sound design quality. We show that this approach preserves agent performance under equivalent designs while remaining highly sensitive to design degradations.

    \item Efficiency-performance correlation: We provide an empirical analysis of the relationship between latent space efficiency and agent behavior. Specifically, we show how the information-theoretic properties of the representation space correlate directly with the agent's ability to utilize sound cues for decision-making.
\end{itemize}

\section{Related Work}
\subsection{Sound design evaluation}
Several studies have investigated the evaluation of sound design in games \cite{10.1109/CoG51982.2022.9893624,9893718,khan2025cog}. One notable example is the DareFightingICE Sound Design Competition \cite{10.1109/CoG51982.2022.9893624}, which introduced both a sound design track and an AI evaluation track. In this competition, a deep RL blind AI agent \cite{9893718}, which plays games using only sound, was employed as an objective evaluator, measuring how effectively different sound designs supported gameplay performance. In addition, Khan et al. \cite{khan2025cog} proposed an evaluation framework using both blind agents and human players for sound design evaluation in a first-person shooter game. While these studies demonstrate the feasibility of agent-based sound design evaluation, our work investigates this problem from a different perspective by leveraging multimodal representation learning and applying it to alternative game environments.

\subsection{Representation learning in reinforcement learning}
Representation learning has been widely explored in RL to address the challenges of high-dimensional sensory input, such as images or audio \cite{hafner2019learning,ha2018world}. Autoencoder-based approaches, including variational autoencoders, have shown that compact latent spaces can improve the sample efficiency and transferability of RL agents. More recent studies have introduced contrastive objectives to further enhance the robustness of learned representations, for example, through predictive coding \cite{oord2018representation} or cross-modal alignment \cite{stooke2021decoupling,lygerakis2024m2curl,becker2023combining}. Nevertheless, state representation has been explored in offline visual RL to isolate task-relevant factors and reduce the impact of spurious correlations \cite{ZHANG2025113565,cao2025towards}. These studies demonstrate that the combination of autoencoding and contrastive methods provides effective representations for downstream RL tasks.

These representation learning techniques are applied distinctly across RL paradigms. In model-based approaches, latent representations are primarily learned to capture environment dynamics and support efficient planning, as demonstrated in methods like World Models \cite{ha2018world}, PlaNet \cite{hafner2019learning}, and the Dreamer series \cite{hafner2019dream,hafner2020mastering,hafner2025mastering}. Conversely, in model-free approaches, representations are typically shaped by auxiliary objectives to improve the feature quality for a policy or value function. Our work builds on these principles, employing a contrastive learning framework to develop representations from sound and vision for the purpose of evaluating game sound design.
\subsection{Multimodal learning}
Multimodal learning seeks to combine information from different sensory inputs into a unified representation that captures cross-modal relationships \cite{ngiam2011multimodal}. More recent approaches, including contrastive methods such as CLIP \cite{radford2021learning}, have shown that aligning modalities in a shared latent space can yield robust and transferable representations. In RL, multimodal representation methods such as CoRAL \cite{becker2023combining} and M2CURL \cite{lygerakis2024m2curl} have been explored for integrating vision with proprioceptive or tactile signals. Not only RL, multimodal learning has also been explored in other tasks such as classification \cite{LI2025113529} and self-supervised learning \cite{zou2025self}.

Fusion strategies in multimodal learning vary from early fusion of raw features to late fusion at the decision level \cite{TsaiLZMS19,tsai2019multimodal,liang2021multibench}, with shared latent representations emerging as a particularly effective approach for exploiting cross-modal relationships. Compared to generative-based methods, fusion methods can be more resilient to imperfections in any single data modality \cite{barnum2020benefits}. If one modality is noisy or incomplete, correlated information from another modality (e.g., the visual channel) can help recover suppressed features, resulting in a more robust and reliable representation \cite{10.1145/3242969.3243014}. Despite this progress, relatively little attention has been devoted to game audio, even though sound is a critical modality for accessibility and immersion in games. Motivated by these advances, our work applies multimodal representation learning to align vision and sound in games, enabling systematic evaluation of sound designs through their effect on agent performance.

\section{Methodology}

In this work, the task of sound design evaluation is formulated as measuring the performance of an AI agent when it relies on audio cues to act in a game. An effective sound design should provide informative cues that enable the agent to achieve higher performance, while degraded or ambiguous designs are expected to reduce performance. We therefore use agent performance as an objective and reproducible metric for comparing different sound designs.

For each game, we adopted a two-step training procedure consisting of pretraining and fine-tuning. In the pretraining phase, we first trained a multimodal encoder using a baseline sound design together with visual input to learn a joint latent representation. A policy network was then trained to take this latent representation as input and output actions for the agent. Algorithm \ref{alg:pretrain} and Algorithm \ref{alg:policy} show the details of pretraining and training policy network, respectively.

The multimodal encoder learns a joint latent representation from both visual and audio information using the baseline sound design. At each game step, three types of observations can be derived from the environment: (i) the \textit{image observation} - the raw game frame; (ii) the \textit{audio observation} - a raw representation of game audio; and (iii) the \textit{audio-image} - a fused representation that combines the audio and image observations into a single multimodal feature vector. The fusion of audio-image is the input to the policy network during policy training, while in the blind evaluation, only the audio observation is provided.

\begin{figure*}[bt]
    \subfloat[Our method pipeline.]{%
         \includegraphics[width=0.5\linewidth]{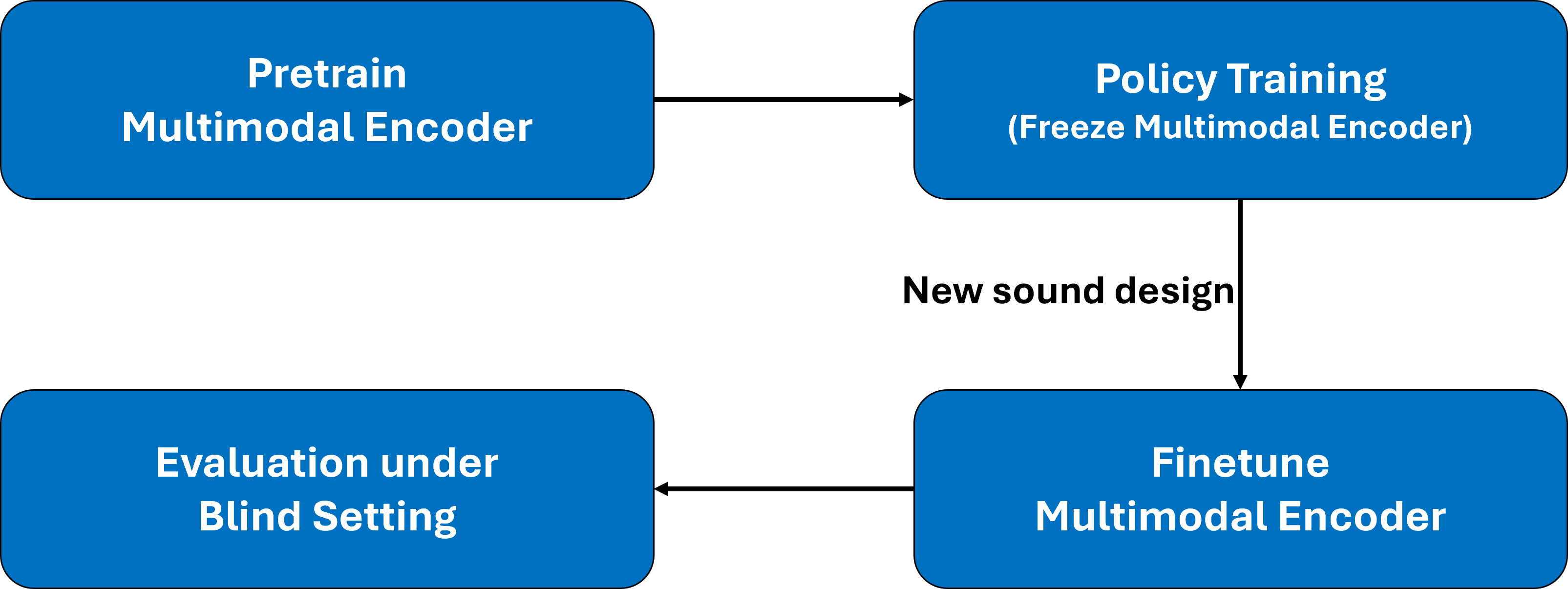}
      }\hfill
    \subfloat[Comparison between evaluation under standard setting and blind setting.]{%
         \includegraphics[width=0.5\linewidth]{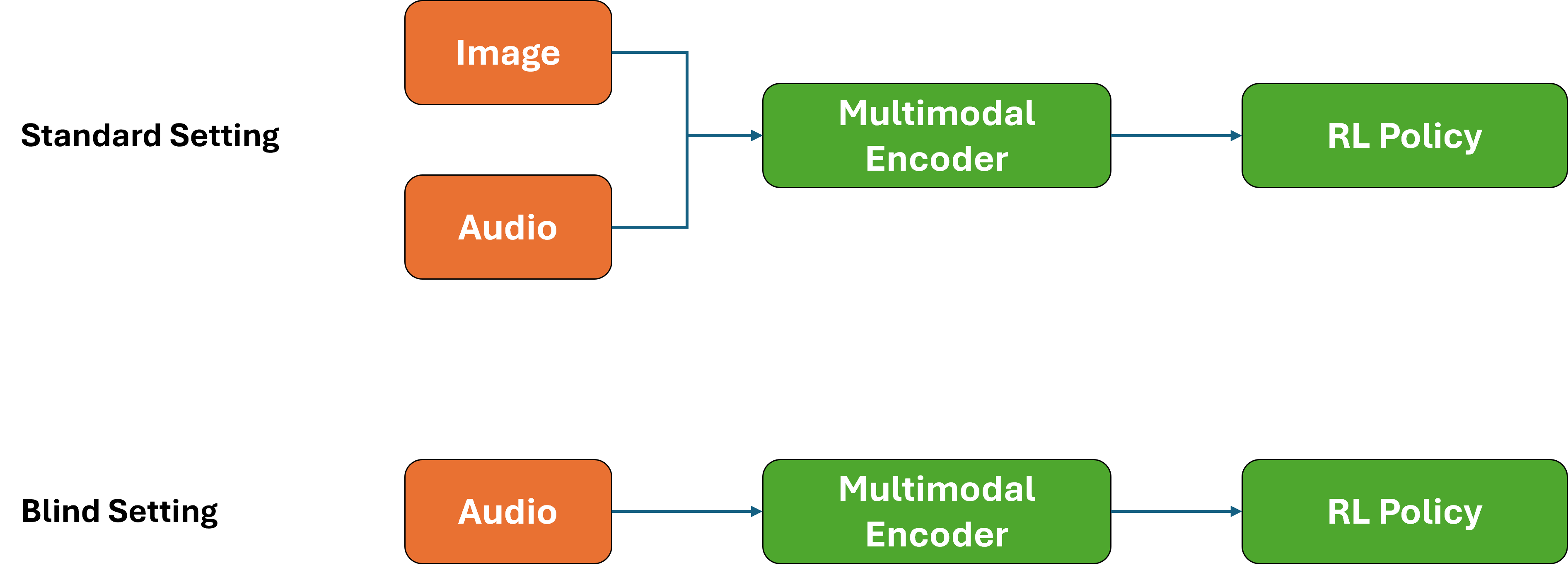}
      }\hfill
    \caption{Our method pipeline and agent evaluation setting. Only audio is used in the blind setting, while both images and audio are used for the standard setting.}
    \label{fig:pipeline_eval}
\end{figure*}

\begin{figure}[bt]
\captionsetup{justification=centering}
\includegraphics[width=0.5\textwidth]{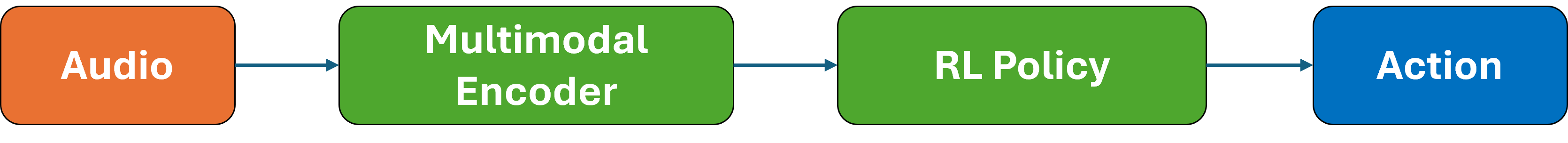}
\caption{Blind Agent Architecture. Audio input is processed by a multimodal encoder to extract latent features, which the RL policy uses to output an action.}
\label{fig:architecture}
\end{figure}

In the fine-tuning stage, only the audio-specific components were updated, while the remaining parts of the multimodal encoder and the policy network were kept frozen. This design allowed us to adapt the representation to new sound design without retraining the policy, thus saving computational cost and ensuring consistency across sound designs. To evaluate performance, we performed zero-shot policy transfer, where the pretrained policy was directly reused with the output of the finetuned encoder. Finally, we measured performance in a \textbf{blind setting}. In the blind setting, the visual modality, which was present during training, is removed exclusively during testing to assess how effectively the agent can perform tasks using only auditory cues. Algorithm \ref{alg:finetune} shows the details of finetuning process. Although it uses the same objective function as Algorithm \ref{alg:pretrain}, it updates only the audio-related components, keeping all other network components frozen. Figure \ref{fig:pipeline_eval} shows our method pipeline and evaluation method and the overall architecture of the agent is shown in Figure \ref{fig:architecture}.

\begin{algorithm}[!tbp]
\caption{Stage 1: Pretrain Multimodal Encoders}
\label{alg:pretrain}
\begin{algorithmic}[1]
\Require Dataset $\mathcal{D}=\{(x^{\mathrm{img}}, x^{\mathrm{aud}})\}$ from the baseline sound design; batch size $B$; temperature $\tau$
\State Initialize encoders $E_{\mathrm{img}}, E_{\mathrm{aud}}, E_{\mathrm{fuse}}$ and common head $C$
\While{not converged}
  \State Sample $\{(x^{\mathrm{img}}_b,x^{\mathrm{aud}}_b)\}_{b=1}^B \sim \mathcal{D}$
  \For{$b \gets 1$ \textbf{to} $B$}
      \State $h^{\mathrm{img}}_b \!\gets\! E_{\mathrm{img}}(x^{\mathrm{img}}_b)$;\quad $h^{\mathrm{aud}}_b \!\gets\! E_{\mathrm{aud}}(x^{\mathrm{aud}}_b)$
      \State $z^{\mathrm{img}}_b \!\gets\! C(h^{\mathrm{img}}_b)$;\quad $z^{\mathrm{aud}}_b \!\gets\! C(h^{\mathrm{aud}}_b)$
      \State $z^{\mathrm{fuse}}_b \!\gets\! C(E_{\mathrm{fuse}}(x^{\mathrm{img}}_b \oplus x^{\mathrm{aud}}_b))$
  \EndFor
\State Compute GMC loss \newline $\mathcal{L}_{\mathrm{GMC}}(\{z^{\mathrm{img}}_b\}_{b=1}^B,\{z^{\mathrm{aud}}_b\}_{b=1}^B, \{z^{\mathrm{fuse}}_b\}_{b=1}^B;\tau)$
  \State Update the parameters of $\{E_{\mathrm{img}},E_{\mathrm{aud}},E_{\mathrm{fuse}},C\}$ by descending $\nabla \mathcal{L}_{\mathrm{GMC}}$
\EndWhile
\State \textbf{return} $E_{\mathrm{img}},E_{\mathrm{aud}},E_{\mathrm{fuse}},C$
\end{algorithmic}
\end{algorithm}

\begin{algorithm}[!tbp]
\caption{Stage 2: Train Policy on Latent (use the Fusion encoder when both modalities are present)}
\label{alg:policy}
\begin{algorithmic}[1]
\Require Frozen $E_{\mathrm{img}},E_{\mathrm{aud}},E_{\mathrm{fuse}},$ and $C$;\; RL algorithm (e.g., PPO)
\State Initialize policy $\pi_\phi$ (and value net $V_\psi$ if used)
\While{RL not converged}
  \State Collect rollouts $\{(o_t^{\mathrm{img}}, o_t^{\mathrm{aud}}, r_t)\}_{t=1}^T$, where $\{(o_t^{\mathrm{img}}, o_t^{\mathrm{aud}}\}$ are visual observation and audio observation at timestep $t$, respectively.
  \For{$t \gets 1$ \textbf{to} $T$}

      \State $z_t \gets C(E_{\mathrm{fuse}}(o_t^{\mathrm{img}} \oplus o_t^{\mathrm{aud}}))$

    \State Sample action $a_t \sim \pi_\phi(\cdot \mid z_t)$
    \State $A \leftarrow A \cup \{a_t\}$
    \State $Z \leftarrow Z \cup \{z_t\}$
    \State $R \leftarrow R \cup \{r_t\}$
  \EndFor
  \State Update $\phi$ (and $\psi$) with the chosen RL objective function $O(A, Z, R)$
\EndWhile
\State \textbf{return} trained policy $\pi_\phi$ (and value net $V_\psi$ if used)
\end{algorithmic}
\end{algorithm}

\begin{algorithm}[!tbp]
\caption{Stage 3: Finetune Audio (and Fusion) for New Sound Design; Policy Reuse}
\label{alg:finetune}
\begin{algorithmic}[1]
\Require New dataset $\mathcal{D}'=\{(x^{\mathrm{img}}, x^{\mathrm{aud}}_{\mathrm{new}})\}$;\; frozen $E_{\mathrm{img}}, C,$ and both the image features block and MLP block of $E_{\mathrm{fuse}}$;\; trainable $E_{\mathrm{aud}} $ and the audio features block of $E_{\mathrm{fuse}}$
\While{not converged}
  \State Sample $\{(x^{\mathrm{img}}_b,x^{\mathrm{aud}}_b)\}_{b=1}^B \sim \mathcal{D}'$
  \For{$b \gets 1$ \textbf{to} $B$}
      \State $z^{\mathrm{img}}_b \gets C(E_{\mathrm{img}}(x^{\mathrm{img}}_b))$ \; 
      \State $z^{\mathrm{aud\_new}}_b \gets C(E_{\mathrm{aud}}(x^{\mathrm{aud}}_{b}))$ \; 
      \State $z^{\mathrm{fuse\_new}}_b \gets C(E_{\mathrm{fuse}}(x^{\mathrm{img}}_b \oplus x^{\mathrm{aud}}_{b}))$ \; 
  \EndFor
\State Compute GMC loss \newline $\mathcal{L}_{\mathrm{GMC}}(\{z^{\mathrm{img}}_b\}_{b=1}^B,\{z^{\mathrm{aud\_new}}_b\}_{b=1}^B, \{z^{\mathrm{fuse\_new}}_b\}_{b=1}^B;\tau)$
  \State Update only $E_{\mathrm{aud}}$ and the audio features block of $E_{\mathrm{fuse}}$ by descending $\nabla \mathcal{L}_{\mathrm{GMC}}$
\EndWhile
\State \textbf{return}  $E_{\mathrm{img}},E_{\mathrm{aud}},E_{\mathrm{fuse}},C$
\end{algorithmic}
\end{algorithm}

\begin{figure*}[t!]
\centering
\captionsetup{justification=centering}
\includegraphics[width=0.70\textwidth]{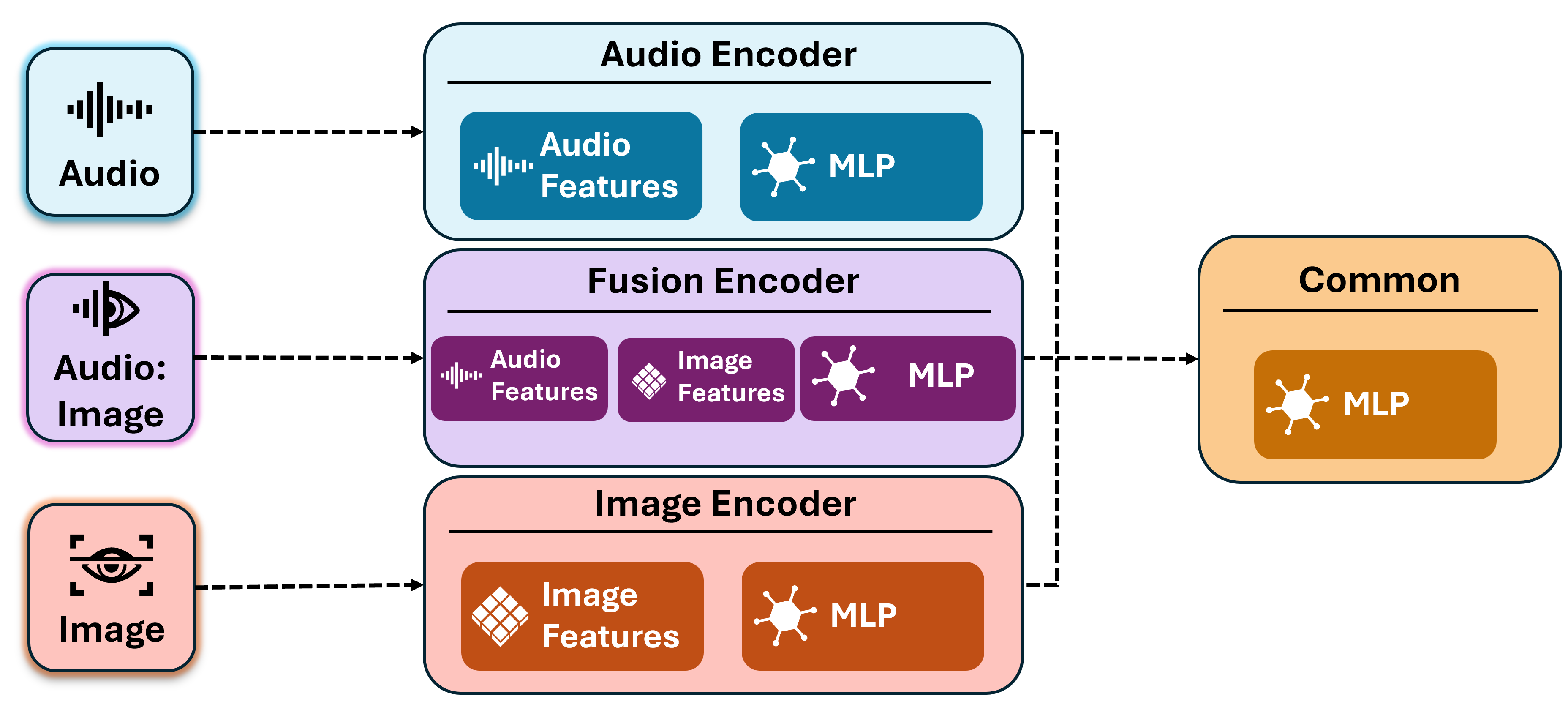}
\caption{Multimodal encoder architecture. Audio, Image, and fused modalities are processed by dedicated feature extractors and MLPs before converging into a shared Common MLP module.}
\label{fig:gmc}
\end{figure*}
\subsection{Multimodal Encoder}

We used the geometric multimodal contrast (GMC) multimodal encoder introduced in \cite{pmlr-v162-poklukar22a} to learn latent representations of the visual and audio modalities because it is lightweight and robust to missing modality. 
For completeness, we provide a brief summary of the model, adapted from the original work. 
An illustration of the multimodal encoder is shown in Figure~\ref{fig:gmc}. 
The GMC multimodal encoder comprises the following components:
\begin{itemize}
    \item Modality-specific encoders that process unimodal inputs and fused representations of the visual and audio modalities, producing intermediate features. There are three encoders: image encoder, audio encoder, and fusion encoder.
    \item A shared projection head, which is a Multilayer Perceptron (MLP), mapping the intermediate features into the latent space.
    \item To train the multimodal encoder, a multimodal contrastive loss function $(\mathcal{L}_{\mathrm{GMC}})$ that aligns the geometric structure of modality-specific representations is used. The definition of this loss function is given later.
\end{itemize}
\subsubsection{Training Objective}
Let $sim(u, v)$ be the cosine similarity between vectors $u$ and $v$, and let $\tau \in (0, \infty)$ be the temperature. The similarity between two representations $z^i_p$ and $z^j_q$ is defined as below:
\begin{equation}
    s_{(p, q)}(i, j) = exp(sim(z^i_p, z^j_q)/\tau), 
\end{equation}
where $z^i_p$ and $z^j_q$ correspond to the $i$th and $j$th samples from a batch $\mathcal{B}=\{z^i_{1:M}\}^B_{i=1}$ which is a batch of $B$ complete representations. For a given modality $m$, $(z_m^i, z^i_{1:M})$ and $(z^i_{1:M}, z^i_m)$ are positive pairs for $i = 1,\dotsc, B$ and the remaining pairs are negative. Therefore
\begin{equation}
    \Omega_{(p,q)}(i) = \sum_{i\neq j}s_{p,p}(i,j) + \sum_{i\neq j}s_{p,q}(i,j)
\end{equation}
is the sum of similarities among negative pairs corresponding to the positive pair $(z^i_p, z^i_q)$, where a positive pair consists of modality representations from the same sample, and a negative pair consists of representations from different samples. The contrastive loss for the positive pairs $(z^i_m, z^i_{1:M})$ and $(z^i_{1:M}, z^i_m)$ is defined as
\begin{equation}
l_m(i) = -\log \frac{s_{m,1:M}(i,i)}{\Omega_{m,1:M}(i)}
         -\log \frac{s_{1:M,m}(i,i)}{\Omega_{1:M,m}(i)}.
\end{equation}
Finally, the loss function for the batch $\mathcal{B}$ is defined as follows.
\begin{equation}
    \mathcal{L}_{\mathrm{GMC}}(\mathcal{B})
   = \sum_{m=1}^{M} \sum_{i=1}^{B} l_m(i).
\end{equation}


\section{Experiments}

We evaluated our framework in three games: PENDULUM, HYPERHOT, and VizDoom, which represent different gameplay genres. In these games, the reward function does not explicitly model perceptual sound quality beyond agent task performance. Rather, the quality of a sound design is defined by its informativeness; designs with rich cues facilitate better decision-making, while degraded designs convey limited or ambiguous information.
For each game, we prepared four sound designs. Design 1 serves as the baseline and is used for multimodal pretraining. Design 2 is semantically equivalent to the baseline and is used in fine-tuning to confirm whether the agent performs consistently under equivalent conditions. Design 3 is deliberately degraded with less informative cues to evaluate the agent's ability to detect quality loss. Finally, Design 4 introduces further degradation compared to Design 3. The details of their conditions for each game are described in the following subsections.

\subsection{PENDULUM}
\begin{figure*}[!tb]
  \centering
  \subfloat[Design 1]{%
     \fbox{\includegraphics[width=0.2\linewidth]{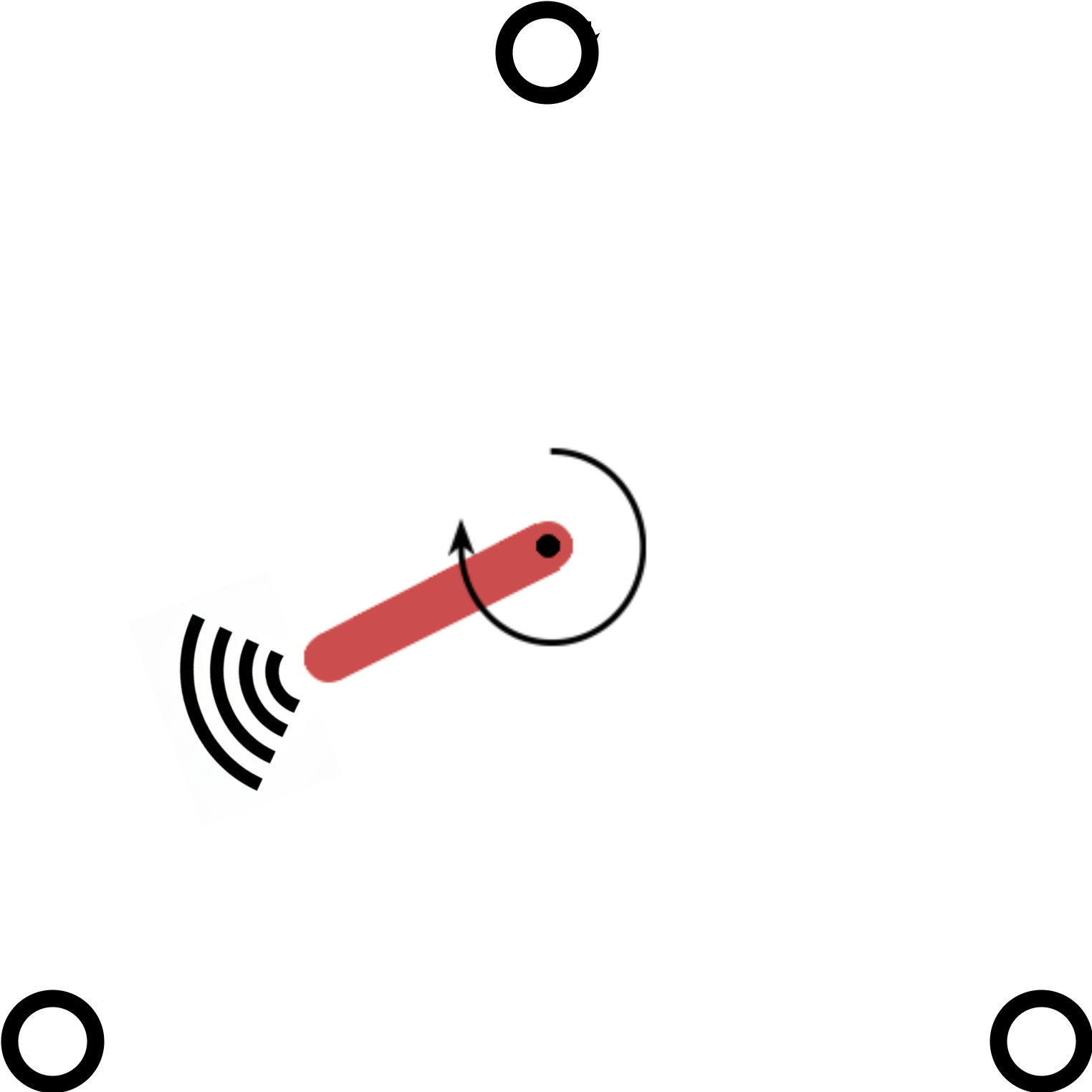}}
  }\hfill
  \subfloat[Design 2]{%
     \fbox{\includegraphics[width=0.2\linewidth]{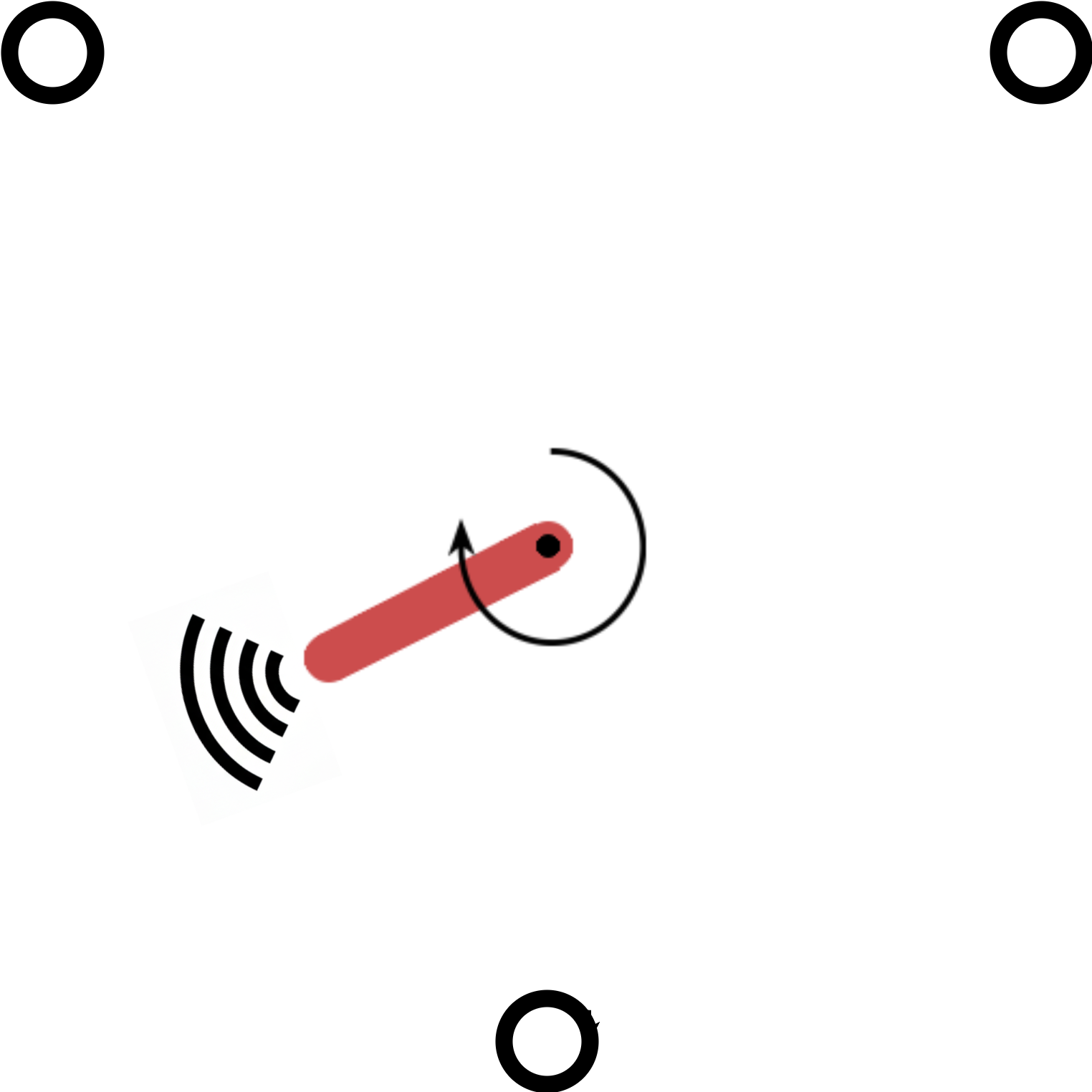}}
  }\hfill
  \subfloat[Design 3]{%
     \fbox{\includegraphics[width=0.2\linewidth]{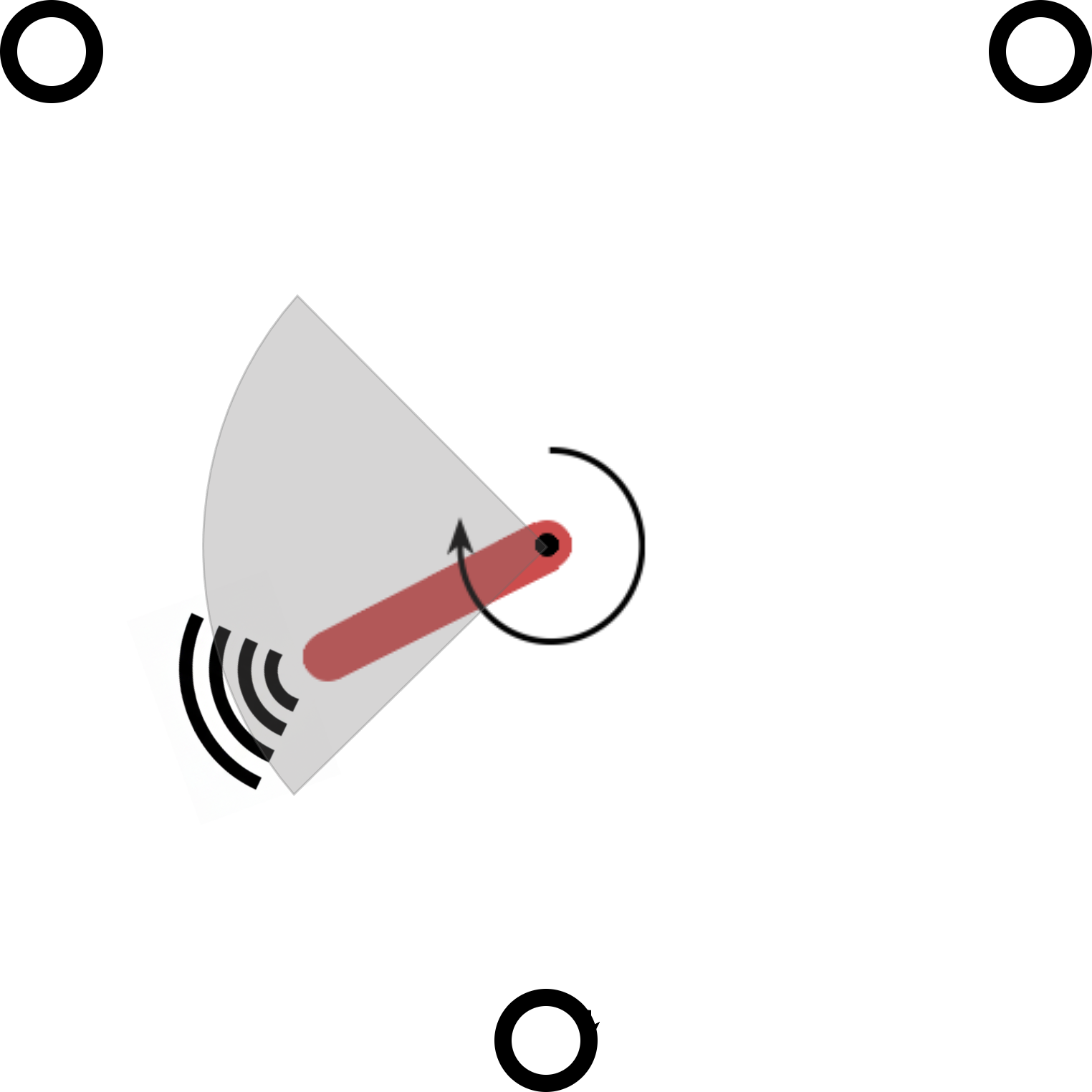}}
  }\hfill
  \subfloat[Design 4]{%
     \fbox{\includegraphics[width=0.2\linewidth]{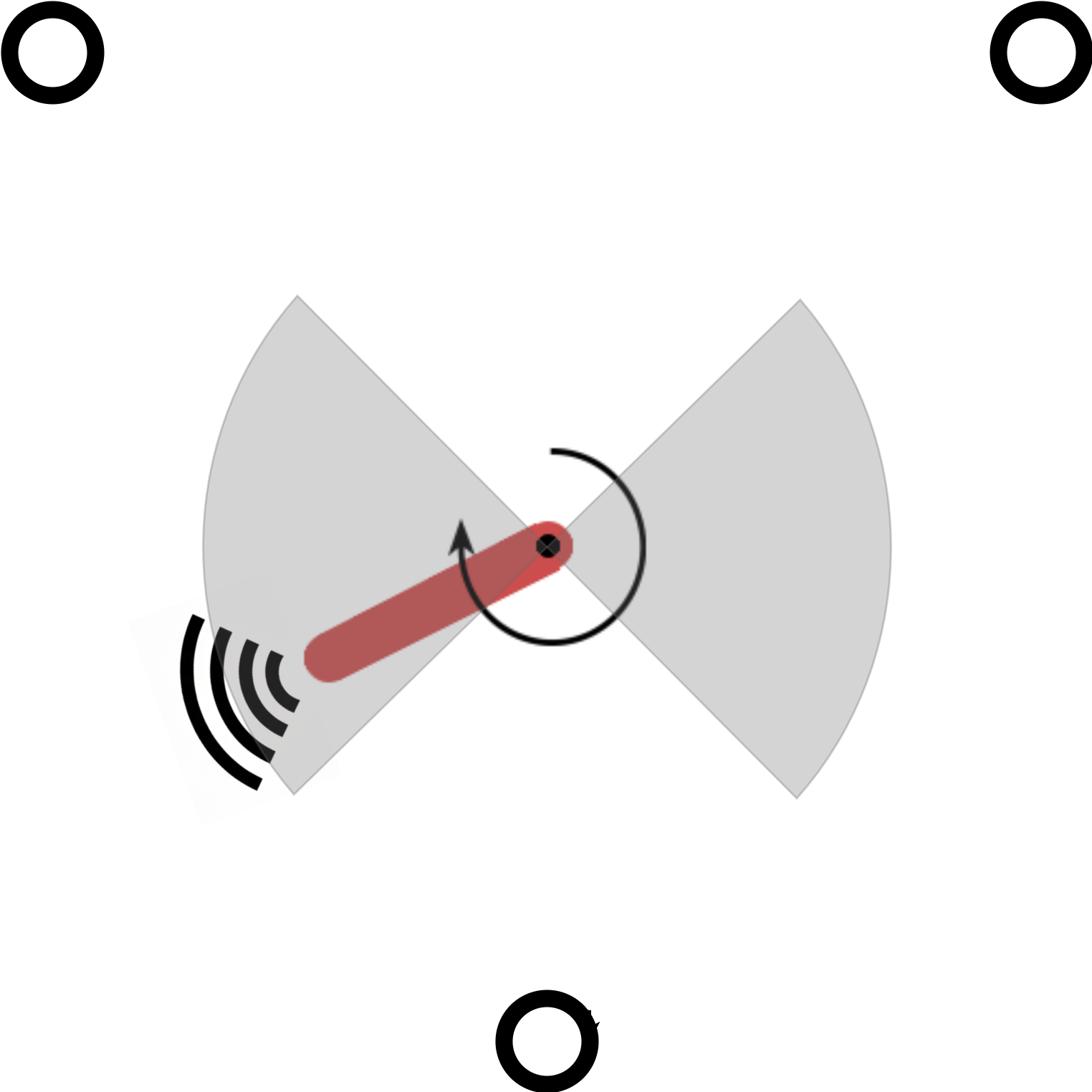}}
  }\hfill
  \caption{PENDULUM sound designs. \newline
  (a) Design 1: three listeners positioned at the bottom left and right, and middle top.\newline
(b) Design 2: same as (a) but the three listeners are moved to the top left and right, and bottom middle.\newline
(c) Design 3: same as (b) but regions where the sound cue is indistinguishable are shaded.\newline
(d) Design 4: same as (b) but regions where the sound cue is indistinguishable are shaded.}
  \label{fig:pendulum}
\end{figure*}

In the PENDULUM game, the task is to swing the pendulum upward and keep it balanced for as long as possible. We used a modified version of the OpenAI Gym environment \cite{10.5555/3398761.3398907} that provides audio feedback to the agent. Figure \ref{fig:pendulum} shows an example of the pendulum environment with its corresponding sound designs.


\subsubsection{Baseline Sound Design}
For the baseline sound design (Design 1), we adopted the configuration introduced in \cite{10.5555/3398761.3398907}. 
In this setup, the tip of the pendulum continuously emits a base frequency $f_0$, which is received by a set of sound receivers $\{ \boldsymbol{\rho}_{1}, \ldots, \boldsymbol{\rho}_{S} \}$. 
At each timestep, the perceived frequency $f^{\prime}_i$ at receiver $\boldsymbol{\rho}_i$ is determined by the Doppler effect, which models the change in observed frequency as the relative distance between the sound source and receiver decreases or increases. 
The resulting frequency $f^{\prime}_i$ is given by:
\[
f'_{i} =
\left(
\frac{c + \boldsymbol{\dot{\rho}}_{i} \cdot (\mathbf{e} - \boldsymbol{\rho}_{i})}
     {c - \mathbf{\dot{e}} \cdot (\boldsymbol{\rho}_{i} - \mathbf{e})}
\right) f_{0},
\]
where $c$ is the speed of sound, the overdot represents the velocity of its vector, and $\mathbf{e}$ is the position of the sound emitter, $\boldsymbol{\rho}_{i}$ and $\mathbf{e}$ being two-dimension vectors. The amplitude $a_i$ heard by receiver $p_i$ follows the inverse square law
\[
a_{i} = \frac{K}{\lVert \mathbf{e} - \boldsymbol{\rho}_{i} \rVert^{2}},
\]
where $K$ is a scaling constant. In the PENDULUM game, the angle $\theta$ of the pendulum is defined as the angular displacement from the upright vertical position. More specifically
\begin{itemize}
    \item $\theta = 0$: The pendulum is pointing straight up (the goal state).
    \item $\theta = 180$ (or $-180$) degree: The pendulum is pointing straight down.
    \item Positive values: Represent counter-clockwise rotation.
    \item Negative values: Represent clockwise rotation.
\end{itemize}
\subsubsection{Finetuning sound designs}
Based on the baseline sound design, we created three variants. The first variant (Design 2) rotates the listeners’ locations while keeping all other settings unchanged. The second variant (Design 3) uses the same listener positions as Design 2 but restricts the sound to a specific angular range of the pendulum. Specifically, when the pendulum’s angle lies in (45, 135), the emitted sound's frequency is fixed, as if the pendulum were at 90 degree. Design 4 introduces an additional constraint, when the angle lies in (-45, -135), the sound's frequency is fixed as if the pendulum were at -90 degree.

\subsection{HYPERHOT}
\begin{figure*}[!tb]
  \centering
  \subfloat[Design 1]{%
     \includegraphics[width=0.2\linewidth]{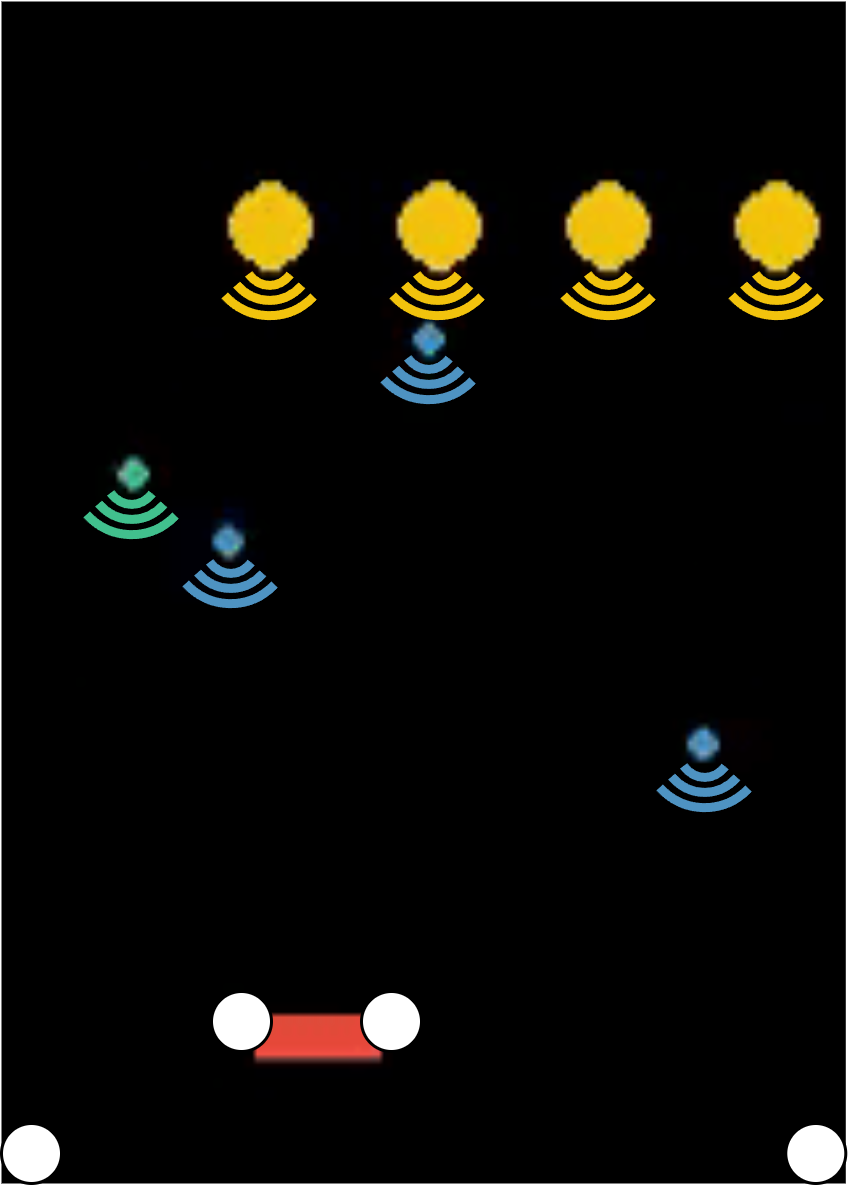}
  }\hfill
  \subfloat[Design 2]{%
     \includegraphics[width=0.2\linewidth]{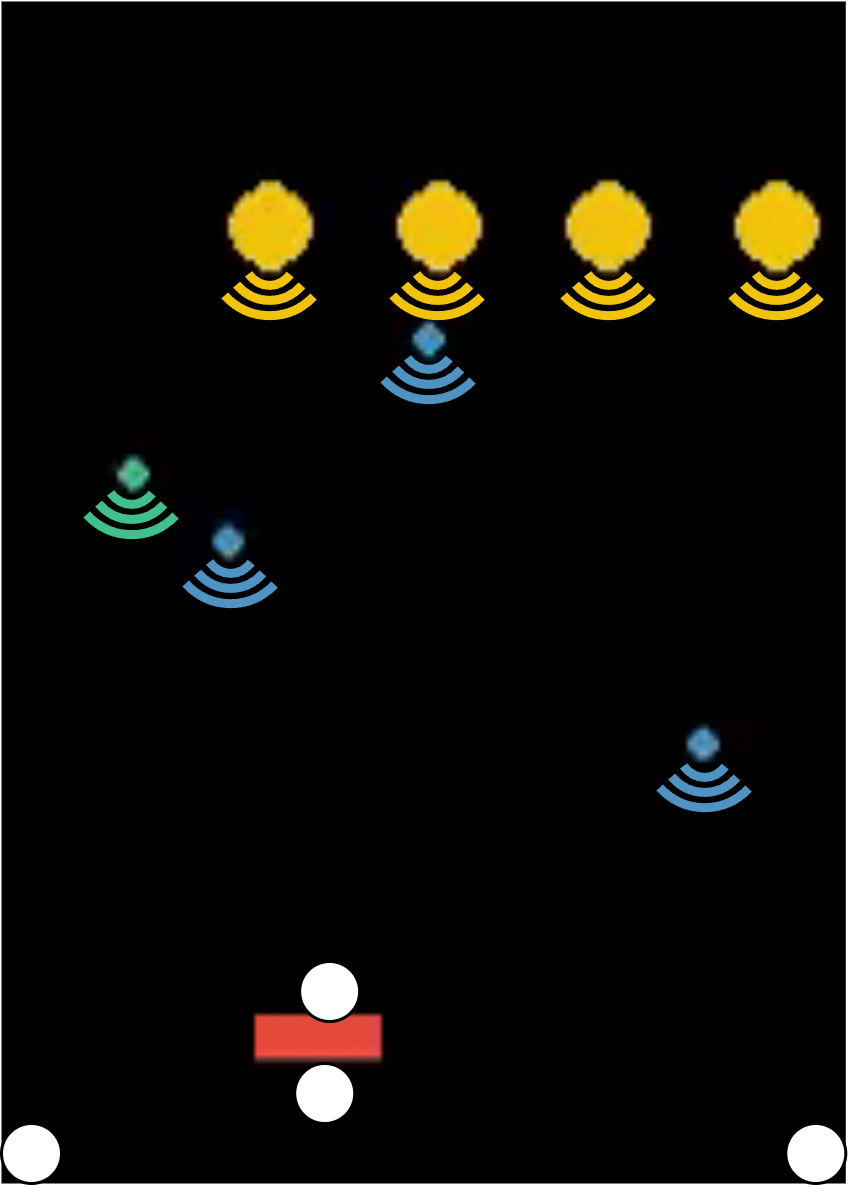}
  }
  \hfill
  \subfloat[Design 3]{%
     \includegraphics[width=0.2\linewidth]{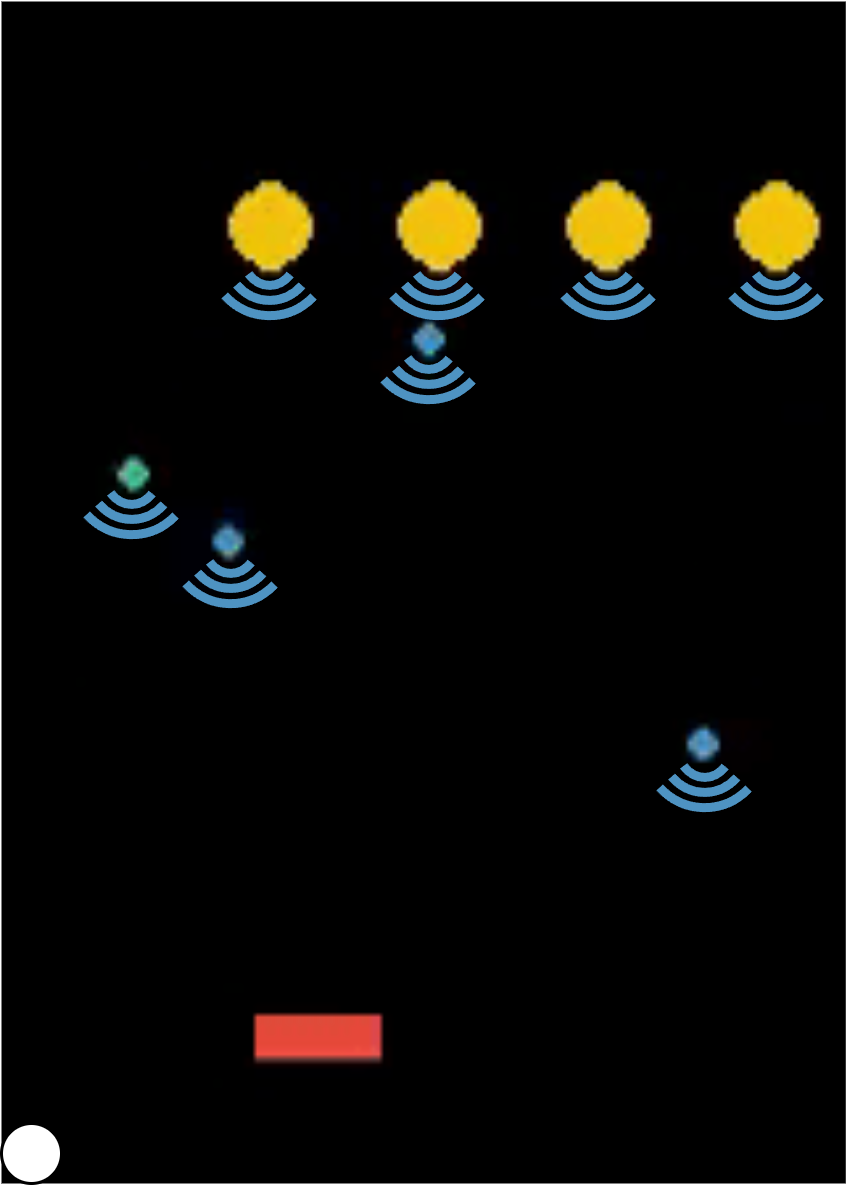}
  }
    \hfill
  \subfloat[Design 4]{%
     \includegraphics[width=0.2\linewidth]{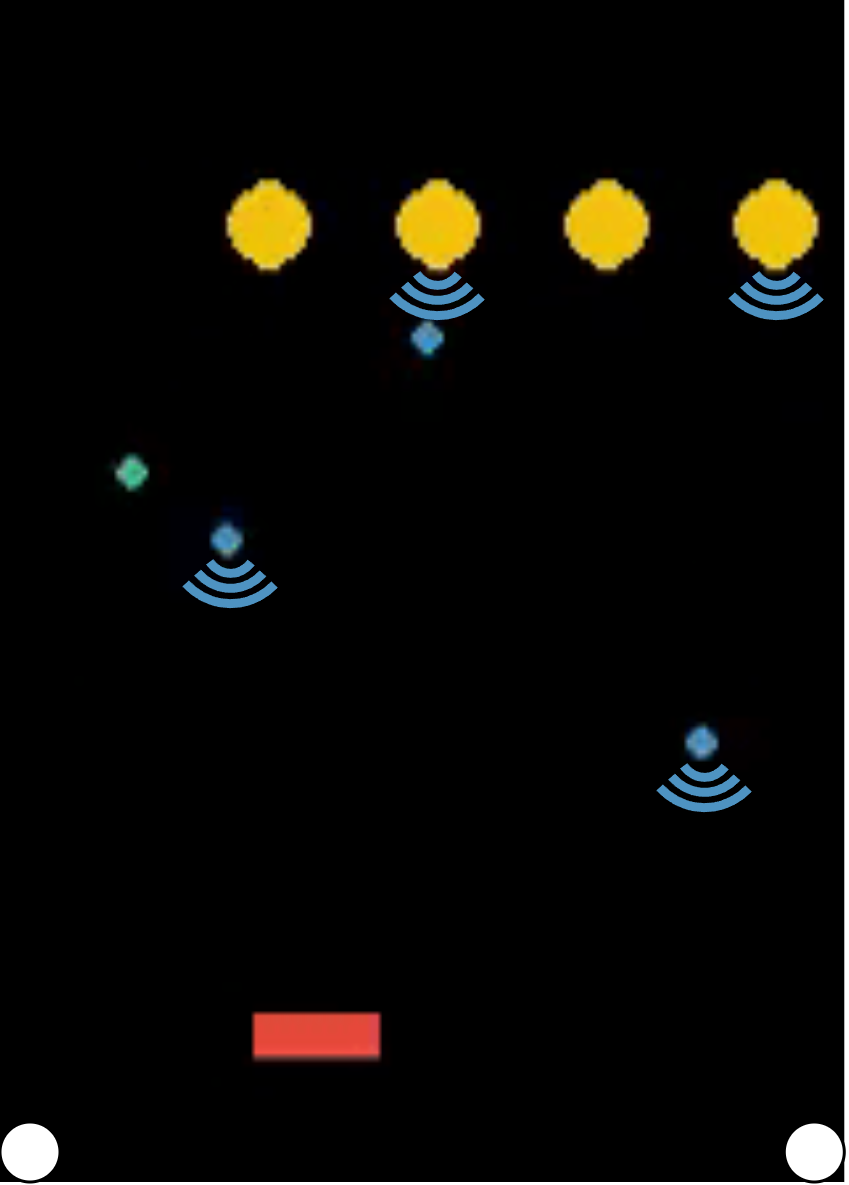}
  }
  \caption{HYPERHOT sound designs. \newline
  (a) Design 1: four listeners (white dots) receive distinct sound cues from enemies (yellow) and bullets (green/blue).\newline
(b) Design 2: same as (a), but listener positions are changed.\newline
(c) Design 3: enemies and bullets emit the same pitch, and only one listener remains at the bottom left.\newline
(d) Design 4: only two enemies and two bullets emit sound, enemies and bullets emit the same pitch, and two listeners remain at the bottom left and bottom right.
}
  \label{fig:hyperhot}
\end{figure*}

HYPERHOT game is a novel top-down shooter game inspired by the SPACE INVADERS Atari game and has been used in previous studies \cite{10.5555/3398761.3398907,10.5555/3535850.3535995}. The objective is to eliminate enemies positioned above while dodging incoming fire by moving horizontally. A sample screenshot of HYPERHOT is provided in Figure~\ref{fig:hyperhot}. The game employs a reward system where the agent gains points for successful actions, defined as follows:
\[
r =
\begin{cases}
10 & \text{if all enemies are killed, \emph{i.e.}, win} \\
-1 & \text{if player is killed or time is up, \emph{i.e.}, lose} \\
0  & \text{otherwise}
\end{cases}
\]

\subsubsection{Baseline sound design}
For the baseline sound design (Design 1), we followed the configuration described in \cite{10.5555/3398761.3398907}. 
In this setup, environmental sound cues are generated by multiple emitters $e_i$, each producing a fixed frequency $f_0^{(i)}$:  
\begin{itemize}
    \item Enemy units on the left ($e_0$) and right ($e_1$) emit sound with frequencies $f_0^{(0)}$ and $f_0^{(1)}$, respectively.  
    \item Enemy projectiles ($e_2$) emit sound with frequency $f_0^{(2)}$.  
    \item The agent’s projectiles ($e_3$) emit sound with frequency $f_0^{(3)}$.  
\end{itemize}  

These sound cues are detected by a set of $S$ receivers, $\{\rho_1,\ldots,\rho_S\}$. 
At each frame, the contribution of every emitter within the audible region is calculated based on its distance to each receiver. 
The sound wave generated by emitter $e_i$ and perceived at receiver $\rho_j$, denoted $a^{(i)}$, is modeled as:
\[
a^{(i)} = a_{0}^{(i)} \exp\left( -\delta \| \mathbf{e}_i - \mathbf{p}_j \|^2 \right),
\]
where $\delta$ is a scaling constant, and $\mathbf{e}_i$ and $\mathbf{p}_j$ denote the positions of emitter $e_i$ and receiver $\rho_j$, respectively.
\subsubsection{Fine-tuning Sound Designs}
We created three sound designs to fine-tune the encoder from the baseline. Design 2 was semantically equivalent to Design 1. However, listeners' locations were changed. For Design 3, we made the sound of enemies and bullets the same, with unchanged amplitude at varying distances, and used only one listener. As a result, Design 3 was less informative than both Design 1 and Design 2. Design 4 further restricts the auditory field by limiting sound emission to only two enemies and their bullets, while preserving the indistinguishability of enemy and bullet sounds and the constant amplitude characteristics of Design 3, but utilizing two listeners.

\subsection{VizDoom}
VizDoom\cite{Kempka2016ViZDoom} is an AI research platform for games based on Doom, a first-person shooter game. It supports the development of game AI agents that play the game using visual and audio buffers, making it suitable for research in deep reinforcement learning. There have been various studies about reinforcement learning conducted, where VizDoom serves as the environment\cite{Wydmuch2019ViZdoom,10.1109/CoG52621.2021.9619096,10.1145/3573428.3573729,9668479}. 

In this study, we utilized the Basic scenario\footnote{\url{https://vizdoom.farama.org/environments/default/\#basic}}
 provided by the VizDoom team. The environment consists of a rectangular map with uniform gray walls, ceiling, and floor. The agent is initialized at the center of one of the longer walls, while a red circular monster is spawned at a random location along the opposite wall. The agent’s action space is restricted to lateral movement (left/right) and shooting. A single successful shot eliminates the monster, and each episode terminates either upon the monster’s elimination or when a predefined time limit is reached. In this scenario, the agent is rewarded for killing the monster, with the following reward function:
 \[
r =
\begin{cases}
106 & \text{for killing the monster} \\
-5 & \text{for every shot} \\
1  & \text{for every tic the agent is alive}
\end{cases}
\]

\subsubsection{Baseline sound design}
For the baseline sound design, we adopted the enhanced configuration proposed by \cite{khan2025cog} (Design 1). This design introduces an auditory cue to indicate the presence of an enemy, as well as a feedback sound that is triggered when the monster is successfully hit, thereby confirming the effectiveness of the action.

\subsubsection{Fine-tuning Sound Designs}
We evaluated three alternative sound designs alongside the baseline. While Design 2 is semantically equivalent to Design 1, it employs a quadratic scaling for the monster’s volume relative to its distance from the agent, whereas Design 1 utilizes linear scaling. Consequently, these designs provide identical information through different delivery methods. Design 3 introduces a spatial constraint, triggering sound only when the enemy appears on the left side of the player's camera. Finally, Design 4 represents the default sound design of the Basic scenario, where there are no sound cues indicating the presence of the enemy. In terms of informational density, Design 4 is semantically weaker than Design 3, which in turn is weaker than both Design 1 and Design 2.
\begin{figure*}[t]
  \centering
  \subfloat[Design 1]{%
     \includegraphics[width=0.20\linewidth]{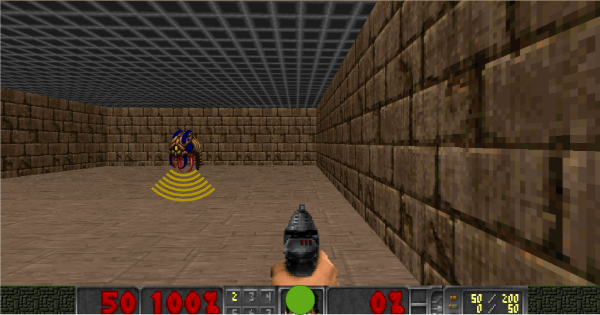}
  }\hfill
  \subfloat[Design 2]{%
     \includegraphics[width=0.20\linewidth]{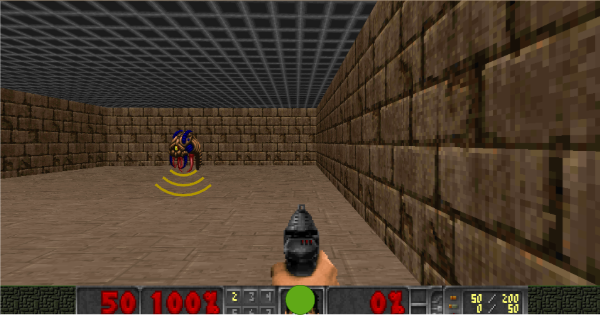}
  }\hfill
  \subfloat[Design 3]{%
     \includegraphics[width=0.20\linewidth]{images/vizdoom_design_2.png}
  }\hfill
  \subfloat[Design 4]{%
     \includegraphics[width=0.20\linewidth]{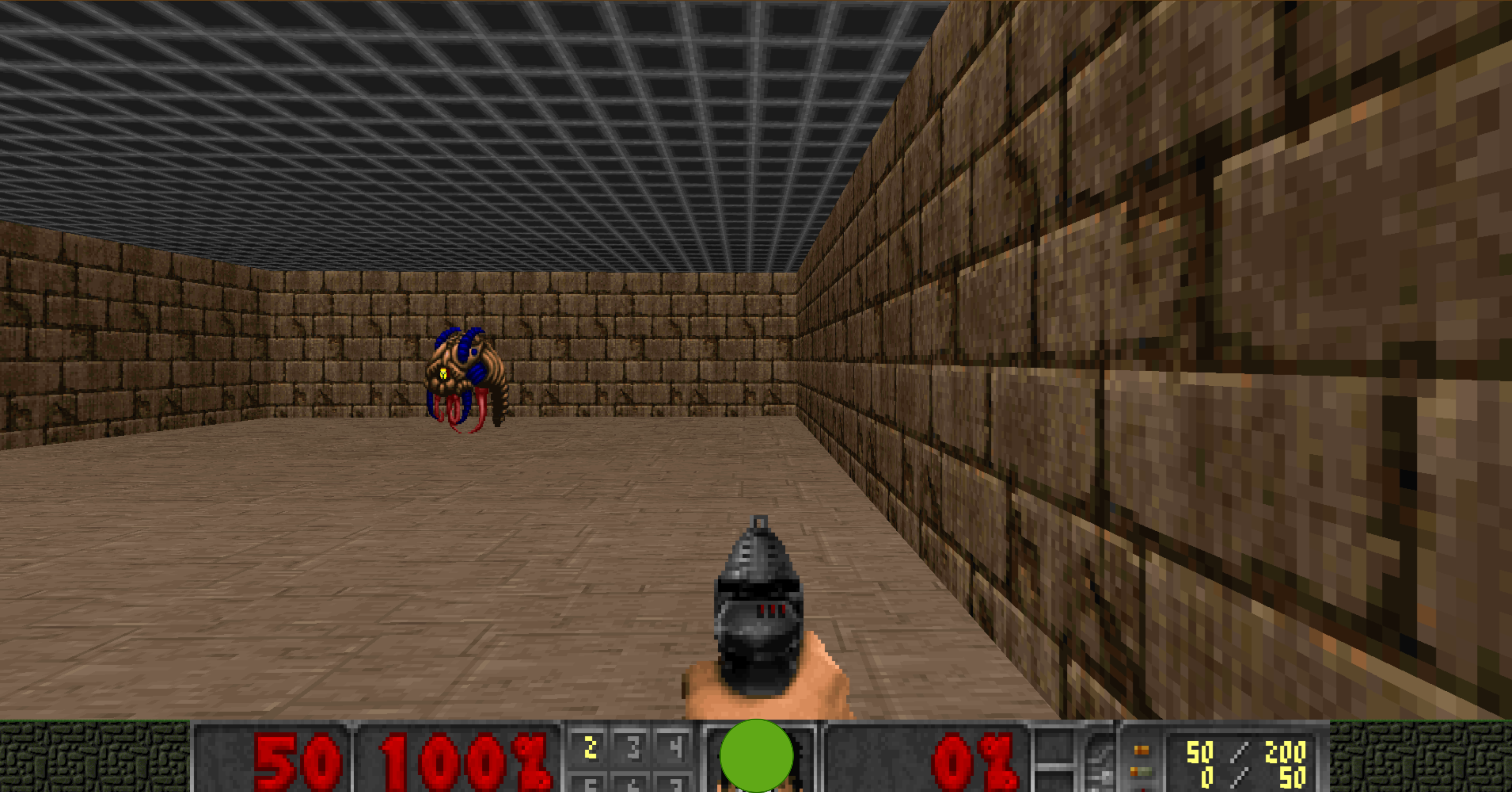}
  }
  \caption{VizDoom basic scenario sound designs. The listener is shown as a green dot at the bottom.\newline
(a) Design 1: enemy emits a sound cue whose volume decreases linearly with distance.\newline
(b) Design 2: same as (a) but the emitted volume decreases quadratically with distance.\newline
(c) Design 3: the player can only hear the enemy's sound if it is on the left side of the player's view.\newline
(d) Design 4: there is no sound cues indicating the presence of enemy.
}
  \label{fig:Vizdoombasic}
\end{figure*}
\section{Results and Discussions}
\subsection{Evaluation Setup and Hyperparameters}
Across all three games, we adopted a consistent evaluation procedure. The multimodal encoder was first pretrained with the baseline sound design and visual input, and a policy network was trained on the resulting latent representation. For each new sound design, the multimodal encoder was finetuned and the policy network was kept intact. The finetuned multimodal encoder then provided input to the trained policy, enabling zero-shot transfer without retraining the agent. Agent performance was evaluated in blind mode using sound-only input, and results were reported as both gameplay performance and computational cost.

For all games, the baseline agent trained from scratch was implemented by replacing the multimodal encoder with a single audio encoder. The hyperparameter and encoder architecture configurations were then adapted for each game. To ensure fair benchmarking and prevent manual tuning bias, we adopted established hyperparameters from prior literature. Specifically, the multimodal encoder training and the Pendulum RL configurations follow \cite{pmlr-v162-poklukar22a}. For Hyperhot and ViZDoom, the RL hyperparameters match those validated by \cite{10.5555/3535850.3535995} and \cite{khan2025cog}, respectively. For the environment, we used a single machine with an A100 NVIDIA GPU. Table T1 in the supplementary document summarizes the hyperparameters used to train the multimodal encoder for the three games. Tables T2, T3, and T4 in the supplementary document present the hyperparameters used to train the policy network and to perform full training from scratch in each game. Figures F1, F2 and F3 in the supplementary document show the detailed architecture of the multimodal encoder in PENDULUM, HYPERHOT, and VizDoom, respectively.

\subsection{Effectiveness of Multimodal Encoder Finetuning}
\Cref{tab:combined_d2_d3_d4} presents the agent's performance across Design 2, Design 3, and Design 4 in the three games. In general, fine-tuning the multimodal encoder leads to distinct performance improvements. The primary exception occurs with Design 4 in ViZDoom. Because this specific design removes crucial audio cues regarding the enemy's location, the encoder cannot extract actionable spatial features, effectively reducing the agent's behavior to random exploration.

Additionally, we evaluated our approach against two baseline methods: training an RL agent from scratch under the blind setting for each sound design, and training a new policy network using the fine-tuned multimodal encoder. Our method achieved comparable performance in Design 2, which shares strong similarities with Design 1. However, performance degraded in Designs 3 and 4. This degradation happens because fine-tuning the multimodal encoder induced a shift in the latent representation space, misaligning it with the frozen policy network's expected inputs. Crucially, this process did not destroy the latent space; when a new policy network was trained on the fine-tuned encoder, the agent's performance consistently improved. This indicates that future research should focus on latent space alignment techniques or incorporate limited policy fine-tuning to accommodate these representational shifts.

\subsection{Agent Performance Across Sound Designs}
The comparative results for all games are summarized in Tables~\ref{tab:combined_performance_validation}, \ref{tab:combined_d2_d3_d4}, and \ref{tab:pairwise_p}. Due to the non-normal distribution of results, we employed the Mann-Whitney U-test and correlation coefficient effect size to evaluate the statistical significance of the difference between two sound designs.  As shown in \Cref{tab:pairwise_p}, because all tests share an identical sample size, we calculated the simple average of the effect sizes for each method. In the following subsections, we analyzed these results for each game in detail.

\subsubsection{PENDULUM}
We evaluated the average reward for each setting over 100 episodes in 10 seeded runs. As shown in Table~\ref{tab:combined_performance_validation}, the agent’s performance with Design 2 was comparable to that with the baseline (Design 1). Furthermore, \Cref{tab:combined_d2_d3_d4} showed that our agent exhibited the same performance trend as both the blind-setting RL agent and the agent trained with a policy network atop the fine-tuned encoder. In all three cases, performance decreased sequentially across Design 2, Design 3, and Design 4.

\begin{table}[tb!]
\centering
\caption{Comparison of finetuned agent performance between the baseline (Design 1) and Design 2 across three games. Statistical significance is evaluated using the Mann-Whitney U test, and the 95\% confidence intervals (CI) for the difference in medians are calculated using a two-sample independent percentile bootstrap method (number of resamples: 9999).}
\resizebox{\columnwidth}{!}{\begin{tabular}{lcccc}
\hline
\textbf{Game} & \textbf{Design 1} & \textbf{Design 2} & \textbf{$p$-value} & \textbf{95\% CI} \\ \hline
Pendulum & -1.020 $\pm$ 0.033 & -1.028 $\pm$ 0.048 & 0.273 & [-0.016, 0.044]\\
Hyperhot & 1.728 $\pm$ 0.183 & 1.780 $\pm$ 0.124 & 0.364 & [-0.299, 0.160]\\
ViZDoom  & 0.798 $\pm$ 0.014 & 0.810 $\pm$ 0.011 & 0.076 & [-0.027, 0.006]\\ \hline
\end{tabular}
}
\label{tab:combined_performance_validation}
\end{table}

\begin{table*}[tb!]
\centering
\caption{Comparison of agent performance across Designs 2, 3, and 4 in different games. No fine-tuning: The multimodal encoder is trained only on Design 1 and is not fine-tuned.}
\begin{tabular}{llccc}
\hline
\textbf{Game} & \textbf{Method} & \textbf{Design 2} & \textbf{Design 3} & \textbf{Design 4}\\ \hline
PENDULUM & Ours & -1.028 $\pm$ 0.048 & -1.382 $\pm$ 0.043 & -2.464 $\pm$ 0.109 \\
 & No fine-tuning & -5.897 $\pm$ 0.298 & -5.807 $\pm$ 0.057 & -5.437 $\pm$ 0.065\\
 & RL training under blind setting & -1.005 $\pm$ 0.091 & -2.236 $\pm$ 0.068 & -2.719 $\pm$ 0.244 \\
 & Finetuned + Policy Training & -1.030 $\pm$ 0.090 & -1.198 $\pm$ 0.047 & -2.298 $\pm$ 0.152  \\ \hline
HYPERHOT & Ours & 1.780 $\pm$ 0.124 & 0.953 $\pm$ 0.088 & 0.904 $\pm$ 0.081\\
& No fine-tuning & 1.701 $\pm$ 0.157 & 0.877 $\pm$ 0.091 & 0.877 $\pm$ 0.091\\
 & RL training under blind setting & 1.93 $\pm$ 0.107 & 1.79 $\pm$ 0.157 & 1.013 $\pm$ 0.074 \\
 & Finetuned + Policy Training & 1.852 $\pm$ 0.124 & 1.697 $\pm$ 0.128 & 1.376 $\pm$ 0.072 \\ \hline
 VizDoom & Ours & 0.810 $\pm$ 0.011 & -0.458 $\pm$ 0.211 & -2.572 $\pm$ 0.190 \\
 & No fine-tuning & 0.805 $\pm$ 0.007 & -1.047 $\pm$ 0.213 & -2.552 $\pm$ 0.258\\
 & RL training under blind setting & 0.800 $\pm$ 0.013 & 0.743 $\pm$ 0.019 & -1.249 $\pm$ 0.273 \\
 & Finetuned + Policy Training & 0.803 $\pm$ 0.008 & -0.105 $\pm$ 0.203 & -1.939 $\pm$ 0.176 \\ \hline
\end{tabular}
\label{tab:combined_d2_d3_d4}
\end{table*}

\begin{table*}[tb!]
\centering
\caption{Pairwise performance comparisons ($p$-values/effect size) using Mann-Whitney U test of sound designs. D2: Design 2, D3: Design 3, D4: Design 4.}
\label{tab:pairwise_p}
\begin{tabular}{llcccc} 
\hline
\textbf{Game} & \textbf{Method} & \textbf{D2 vs D3} & \textbf{D3 vs D4} & \textbf{D2 vs D4} & \textbf{Average Effect size} \\ \hline
PENDULUM & Ours & $1.8 \times 10^{-4}/0.84$ & $1.8 \times 10^{-4}/0.84$ & $1.8 \times 10^{-4}/0.84$ & $0.84$ \\
     & RL training under blind setting &   $1.8 \times 10^{-4}/0.84$  &  $3.3 \times 10^{-4}/0.80$   &   $1.8 \times 10^{-4}/0.84$  & $0.83$ \\
     & Finetuned + Policy Training &   $4.4 \times 10^{-4}/0.79$  & $1.8 \times 10^{-4}/0.84$   &   $1.8 \times 10^{-4}/0.84$ & $0.82$  \\ \hline
HYPERHOT  & Ours  & $1.7 \times 10^{-4}/0.84$  & 0.29/0.24  & $1.8 \times 10^{-4}/0.84$ & $0.64$\\
      &  RL training under blind setting   &   $4.5 \times 10^{-2}/0.44$ & $1.8 \times 10^{-4}/0.84$   & $1.8 \times 10^{-4}/0.84$ & $0.70$\\
      &   Finetuned + Policy Training    &   $1.9 \times 10^{-2}/0.52$    &  $2.4 \times 10^{-4}/0.82$   & $1.8 \times 10^{-4}/0.84$ & $0.73$\\ \hline
VizDoom & Ours & $1.8 \times 10^{-4}/0.84$ & $1.8 \times 10^{-4}/0.84$ & $1.8 \times 10^{-4}/0.84$ & $0.84$\\
      &  RL training under blind setting &   $1.8 \times 10^{-4}/0.84$    &   $1.8 \times 10^{-4}/0.84$    &   $1.8 \times 10^{-4}/0.84$ & $0.84$\\
      &   Finetuned + Policy Training &   $1.8 \times 10^{-4}/0.84$    &   $1.8 \times 10^{-4}/0.84$    &  $1.8 \times 10^{-4}/0.84$ & $0.84$ \\ \hline
\end{tabular}
\end{table*}




\subsubsection{HYPERHOT}
Due to the sparse reward structure in the HYPERHOT game, we evaluated performance by reporting the average number of successful enemy hits per game. For each setting, we conducted 100 episodes across 10 seeded runs. As shown in Table~\ref{tab:combined_performance_validation}, the agent’s performance with Design 2 was statistically indistinguishable from that of the baseline (Design 1). This finding supports our hypothesis that Design 2 conveyed equivalent information to that of the baseline, resulting in comparable agent performance across the two designs. As shown in Table ~\ref{tab:pairwise_p}, although our agent did not show a statistically significant difference in performance between Design 3 and Design 4 ($p = 0.29$), it distinguished between Design 2 and Design 3 more pronouncedly than the other approaches. Furthermore, the agent trained by our approach produced the same overall performance trend as both the blind-setting RL agent and the two-stage fine-tuning method. Across all three approaches, the performance consistently decreased from Design 2, to Design 3, and Design 4.

\subsubsection{VizDoom}
In VizDoom, we evaluated the average reward for each setting over 100 episodes across 10 seeded runs. As shown in Table \ref{tab:combined_performance_validation}, the agent's performance with Design 2 was comparable to that with the baseline (Design 1). 
Furthermore, the agent trained by our approach exhibited the same performance trend as both the blind-setting RL agent and the agent trained with a policy network atop the fine-tuned encoder, as shown in Table~\ref{tab:combined_d2_d3_d4}. In all three of these cases, the performance decreased sequentially across Design 2, Design 3, and Design 4.

Overall, our agent successfully distinguished the differences between sound designs across all three games, aligning with the performance of the other methods. As shown in \Cref{tab:pairwise_p}, the agent trained by our approach achieved the highest average effect size (0.837) in Pendulum and comparable results in ViZDoom. In addition, as shown in \Cref{tab:combined_d2_d3_d4}, compared to the two baseline methods, our agent performed comparably on Design 2, though its performance declined on Design 3 and Design 4. Nevertheless, it could still distinguish between these sound designs just as effectively as the baselines. We attribute this to a practical trade-off between computational resources and overall performance.







\subsection{Computational Resource Comparison}

In addition to agent performance, we evaluated the efficiency of our method using hardware-agnostic metrics for parameter, computational efficiency, and sample efficiency, while excluding hardware-dependent measures such as wall-clock time.
\subsubsection{Parameter Efficiency and Sample Efficiency}
To compare the two approaches, we evaluated both parameter efficiency and sample efficiency. Parameter efficiency was quantified by the number of trainable parameters, which reflected the storage requirements and the memory usage of the optimizer. Sample efficiency was measured by the number of training samples consumed by each approach, providing a direct indication of computational cost.
\subsubsection{Computational Efficiency}
We analyzed computational efficiency by reporting the total training cost in terms of floating-point operations (FLOPs) per sound design. For each method, the cost was estimated by first measuring the FLOPs per training step using \texttt{torch.utils.flop\_counter}, and then multiplying this value by the total number of training steps $S$.
 \[
    Cost_{total} = FLOPs_{step} \times S,
\]
where $FLOPs_{step}$ is backward $FLOPs$ per step. We excluded the computational cost of inference as it was negligible compared to the training cost, and we also excluded the policy network training cost because it was performed only once on the baseline sound design. For each new sound design, only the multimodal encoder was fine-tuned. For the approach involving a policy network trained atop the fine-tuned encoder, the total computational overhead is the aggregate of the multimodal fine-tuning phase and the subsequent reinforcement learning stage.

Tables~\ref{tab:resource_all_games} showed that our approach was substantially more sample-efficient and required fewer trainable parameters than the others. For PENDULUM, the backward FLOPs were higher in our method. Although the common encoder and the MLP block connecting the audio features block and the image features block of the fusion encoder were frozen, their output gradients still had to be computed to propagate the loss signal back to the trainable audio encoder and the trainable audio features block of the fusion encoder. In addition, only in the PENDULUM setting did the output dimension of the image features block (14,400) exceed that of the audio features block (50), whereas in the other games these dimensions were aligned. This further amplified the backward computational load in PENDULUM. As a result, the backward pass incurred additional cost, even though no parameters were updated within the frozen modules.

By comparison, the RL agent trained under the blind setting required the agent to learn by trial-and-error through interaction with the environment, and each sample must be generated via simulation. This process was computationally expensive and often provides a limited learning signal, particularly in tasks with sparse rewards. Furthermore, methods that train a policy network on top of a fine-tuned multimodal encoder incur the costs of both fine-tuning and policy training. In contrast, our method required only a much smaller data set to train the multi-modal representation, thus reducing both the sample requirements and the overall computational cost.
\begin{table*}[tb!]
    \centering
    \caption{Comparison of training cost in PENDULUM, HYPERHOT and Vizdoom.}
    \label{tab:resource_all_games}
    \begin{tabular}{ll ccccc}
         \hline
         \textbf{Game} & \textbf{Method} & \textbf{Samples} & \textbf{Trainable Parameters} & \textbf{Backward FLOPs} & \textbf{Training Steps} & \textbf{Total FLOPs}\\
         \hline
        PENDULUM & Ours & \textbf{20,000} & \textbf{9,664} & 0.25 GFLOP  & \textbf{70,499} & \textbf{17 TFLOP} \\ 
        & RL training under blind setting & 150{,}000 & 340{,}268 & \textbf{0.19 GFLOP} & 150{,}000 & 29 TFLOP \\ 
        & Finetuned + Policy Training & \makecell{20,000 + \\150{,}000} & \makecell{9{,}664 + \\275{,}972} & \makecell{0.25 GFLOP + \\0.36 GFLOP} & \makecell{70{,}499 + \\150{,}000} & \makecell{17 TFLOP + \\54 TFLOP} \\ \hline
        HYPERHOT & Ours & \textbf{32{,}000} & \textbf{9{,}140{,}288} & \textbf{2.58 GFLOP}  & \textbf{56{,}250} & \textbf{145 TFLOP} \\ 
        & RL training under blind setting & 2{,}000{,}000 & 9{,}248{,}220 & 4.8 GFLOP & 2{,}000{,}000 & 9632 TFLOP \\ 
        & Finetuned + Policy Training & \makecell{32{,}000 + \\2{,}000{,}000} & \makecell{9{,}248{,}220 + \\48{,}140} & \makecell{2.58 GFLOP + \\ 0.024 GFLOP} & \makecell{56{,}250 + \\2{,}000{,}000} & \makecell{145 TFLOP + \\ 48 TFLOP} \\ \hline
        VizDoom & Ours & \textbf{32{,}000} & \textbf{245{,}312} & \textbf{0.24 GFLOP} & \textbf{56{,}250} & \textbf{13.5 TFLOP} \\ 
        & RL training under blind setting & $5 \times 10^8$ & 1{,}684{,}741 & 2.3 GFLOP & $5\times 10^8$ & $1.15\times 10^6$ TFLOP \\
        & Finetuned + Policy Training & \makecell{32{,}000 + \\$5 \times 10^8$} & \makecell{245{,}312 + \\853{,}509} & \makecell{0.24 GFLOP + \\18.95 GFLOP} & \makecell{56{,}250 + \\$5\times 10^8$} & \makecell{13.5 TFLOP + \\$9.47 \times 10 ^ 6$ TFLOP} \\ \hline
    \end{tabular}
\end{table*}
\begin{table*}[!tb]
    \caption{Latent representation similarities across sound designs in PENDULUM, HYPERHOT and VizDoom.}
    \centering
    \begin{tabular}{lccccc}
    \hline
    \textbf{Game} & \textbf{Sound Design} & \textbf{Image-Sound} & \textbf{Image-Joint} & \textbf{Sound-Joint} & \textbf{Average}\\
    \hline
    PENDULUM & Design 1 & 1.00 & 1.00 & 1.00 & 1.00\\
            & Design 2 & 0.99 & 1.00 & 0.99 & 0.99\\
            & Design 3 & 0.83 & 0.98 & 0.87 & 0.89\\
            & Design 4 & 0.80 & 0.98 & 0.85 & 0.88\\
    \hline
    HYPERHOT & Design 1 & 0.68 & 0.44 & 0.52  & 0.55\\
    & Design 2 & 0.64 & 0.45 & 0.50 & 0.53\\
    & Design 3 & 0.15 & 0.14 & 0.18 & 0.16\\
    & Design 4 & 0.12 & 0.14 & 0.17 & 0.14\\
    \hline
    VizDoom & Design 1 & 0.49 & 0.53 & 0.97 & 0.66\\
            & Design 2 & 0.52 & 0.62 & 0.93 & 0.69\\
            & Design 3 & 0.33 & 0.53 & 0.85 & 0.57\\
            & Design 4 & 0.08 & 0.44 & 0.70 & 0.41\\
    \hline
    \end{tabular}
    \label{tab:modality_alignment_combined}
\end{table*}
\subsection{Impact of Latent Representation Adaptation on Reinforcement Learning Policy}
\Cref{tab:modality_alignment_combined} presents the average cosine similarity of the latent modality representations across the three games evaluated in our experiments. Each modality was first processed by a modality-specific encoder and then projected into a shared latent space via a common encoder. In all the games, Design 1 and Design 2 showed comparable modality similarities, both in theory and in practice. Since they were constructed to preserve a similar structure, their empirical similarity scores align perfectly with our initial design hypotheses.\par
In PENDULUM, we observed an almost perfect alignment between sound and image, with similarity scores approaching 1.0 for both Design 1 and Design 2. This indicated that the auditory cues in such settings directly captured the underlying game state. However, in more complex environments like HYPERHOT and VizDoom, the similarity values decreased substantially, reflecting the increased difficulty of aligning audio and visual modalities when the dynamics and sound effects are more intricate. While we observe that cross-modality similarity naturally decreases as game scenarios become more complex, this does not preclude the framework's real-world applicability but shifts how the framework is deployed. In highly complex, modern game environments, purely zero-shot transfer with a strictly frozen policy may reach a performance ceiling. However, our framework can serve as a foundation for a semi-automated evaluation pipeline. By allowing for limited, few-shot finetuning of the policy network on top of the adapted multimodal encoder, the agent can rapidly align with complex new audio cues. To validate this hypothesis, we conducted an ablation study by fine-tuning the policy network for only 20\% of the maximum RL training steps. As shown in \Cref{tab:finetune_policy}, this limited adaptation consistently improved agent performance across all sound designs and games, confirming the efficacy of few-shot policy recovery. Crucially, this approach remains more computationally efficient than the standard baseline of training an RL agent completely from scratch for every sound design iteration. Therefore, the proposed method provides a highly scalable solution for iterative audio quality assurance in realistic game development cycles.
\begin{table}[tb!]
    \caption{Performance of fine-tuned policy agents using Design 3 and Design 4 across three games. Values in parentheses denote the performance improvement relative to baseline agents where only the multimodal encoder was fine-tuned}
    \centering
    
    \begin{tabular}{lcc}
        \hline
        \textbf{Game} & Design 3 & Design 4 \\\hline
        Pendulum & -1.081 (+0.301) & -1.893 (+0.571) \\
        Hyperhot & 1.538 (+0.585) & 1.073 (+0.169) \\
        VizDoom & 0.272 (+0.73) & -1.631 (+0.941)\\
        \hline
    \end{tabular}
    \label{tab:finetune_policy}
\end{table}

In contrast, Design 3 and Design 4 represented a poorer design choice compared to Designs 1 and 2, and this was consistently reflected in the marked decline of modality similarity scores across all three games. This reduction in alignment was consistently mirrored by degraded agent performance, showing that the pretrained policy could not fully interpret the altered sound cues without fine-tuning. Taken together, these findings demonstrate that, across three different games, the expected similarity between Design 1 and Design 2 was confirmed both empirically in modality alignment and in the comparable performance of agents. While further validation is required, these results suggest a potential link between cross-modal similarity scores and agent robustness in unseen sound conditions.

\subsection{Limitations}

Our approach has several limitations that point to directions for future research. First and most importantly, our framework evaluates the functional informativeness of sound rather than its perceptual quality or emotional impact. While our AI-based metrics provide an objective measure of sound utility, they do not capture the aesthetic or subjective experience of a human listener. Consequently, this framework is intended to complement, rather than replace, user-centered studies that assess the emotional resonance and artistic value of sound design.

Second, although current evaluation environments are relatively simplified, the framework’s modular architecture is designed to scale to complex, realistic games. By replacing lightweight encoders with advanced audio foundation models and incorporating 3D spatial features, the system can be adapted for high-fidelity titles and partial observability. Third, the transition from from-scratch training to fine-tuning introduces a latent-shift mismatch between the updated encoder and the fixed policy network. Future work should investigate limited policy adaptation or fixed-space constraints to mitigate this. Furthermore, the granular contribution of individual sound cues remains an open question that warrants deeper investigation.

Beyond its theoretical contributions, this framework offers immediate practical utility for game developers. It serves as an objective, automated quality assurance tool that can be integrated into early-stage development pipelines to identify uninformative audio cues. By providing a scalable proxy for sound utility, designers can rapidly iterate on audio-visual balance before committing to expensive and time-consuming human playtesting.

\section{Conclusion}
This study introduced a framework for evaluating in-game sound designs by combining multimodal representation learning with RL. In our approach, only the multimodal encoder was fine-tuned for each new sound design, while the policy network remained fixed. Experimental results showed that the framework provides a highly effective and efficient alternative to training a new agent from scratch for distinguishing differences between sound designs, substantially lowering computational cost and thereby improving scalability. Furthermore, the observed correlation between latent representation efficiency and agent performance highlighted the potential of this approach and points to future research on advancing multimodal learning techniques for sound design evaluation in complex game environments.

\bibliographystyle{IEEEtran}
\bibliography{references}

\vfill
\end{document}